# 基于知识图谱的直线型建筑物模式识别方法

魏智威 [1,2], 肖屹 [3,4], 童莹 [3], 许文嘉 [5], 王洋 [1,2]

1.中国科学院网络信息体系技术重点实验室，北京 100830；2.中国科学院空天信息创新研究院，北京 100830；3.武汉大学资源与环境科学学院，湖北武汉 430079；4.深圳信息职业技术学院软件学院，广东深圳 518172；5.北京邮电大学信息与通信工程学院，北京 100876

## Linear building pattern recognition via spatial knowledge graph

1. Key Laboratory of Network Information System Technology, Institute of Electronic, Chinese Academy of Sciences, Beijing 100830; 2. The Aerospace Information Research Institute, Chinese Academic of Sciences, Beijing 100830; 3. School of Resources and Environment Science, Wuhan University, Wuhan 430079; 4. School of Software Engineering, Shenzhen Institute of Information Technology, Guangdong Shenzhen 518172; 5.School of Information and Communication Engineering, Beijing University of Posts and Telecommunications, Beijing 100876.

**Abstract**: Building patterns are important urban structures that reflect the effect of the urban material and social-economic on a region. Previous researches are mostly based on the graph isomorphism method and use rules to recognize building patterns, which are not efficient. The knowledge graph uses the graph to model the relationship between entities, and specific subgraph patterns can be efficiently obtained by using relevant reasoning tools. Thus, we try to apply the knowledge graph to recognize linear building patterns. First, we use the property graph to express the spatial relations in proximity, similar and linear arrangement between buildings; secondly, the rules of linear pattern recognition are expressed as the rules of knowledge graph reasoning; finally, the linear building patterns are recognized by using the rule-based reasoning in the built knowledge graph. The experimental results on a dataset containing 1289 buildings show that the method in this paper can achieve the same precision and recall as the existing methods; meanwhile, the recognition efficiency is improved by 5.98 times.

**Foundation support**: The National Natural Science Foundation of China (No.41871378); The Youth Innovation Promotion Association Foundation of Chinese Academic of Sciences (No. Y9C0060)

**摘要**：建筑物空间分布模式是物质和社会经济功能等综合作用于地域形成的重要空间结构，以往研究多基于图同构方法利用规则识别分布模式，识别效率不高。知识图谱利用图模型表达实体和实体间关系，并支持高效推理获取图谱中

特定子图模式。因此，本文试图结合知识图谱高效识别直线型建筑物模式：首先，利用属性图数据模型构建表达建筑物间邻近、相似和沿直线排列空间关系的知识图谱；其次，将直线排列识别的结构化规则表达为知识图谱推理的规则，基于规则推理识别直线模式。实验结果表明，本文方法能实现和已有方法相同的准确率和召回率，在包含 1286 个建筑物的数据集上识别效率能提高 5.98 倍。

**关键词**：空间分布；建筑物；知识图谱；空间推理；格式塔原则

**中图分类号**：P208    **文献标示码**：A

# 0. 引言

建筑物是社会经济、人文和自然等因素综合作用于地域的空间反映，会在群组结构上呈现出明显的模式特征[1]，自动提取建筑物模式是城市空间理解[2,3]、数字地图综合[4-6]等领域研究的热点和难点问题。

直线型模式是一种典型的空间分布模式，是其它建筑物群模式识别的基础。依据直线模式的定义策略，现有研究大致可以分为三类：模板匹配方法、邻近图方法和机器学习方法[8-19]。模板匹配方法是将直线模式形式化为模板，通过匹配定义的模板识别模式，如文献[8, 9]将直线模式划分为层次结构，通过形式化不同层次上的空间关系识别直线模式；考虑到建筑物局部异质性明显但整体呈直线分布的现象，文献[10]则提出了基于模板匹配的组合直线模式识别方法。邻近图方法是利用邻近图表达建筑物属性及其相互关系，并结合结构化规则进行图遍历识别模式，如文献[11]依据模式认知规律定义规则对 MST(Minimum Spanning Tree，最小生成树邻近图)剪枝提取直线型排列和曲线型排列；文献[12]利用 stroke 规则探测线性模式；考虑到直线模式的连通性，文献[13, 14]提出了基于邻近图剪枝的多连通直线模式识别方法；考虑人类视觉认知的完形规则，文献[15, 16]提出了结合邻近图剪枝和图形分解的直线模式识别方法；另外，文献[17, 18]证明了 RNG(Relative Neighborhood Graph，相对邻近图)识别线性模式最有效。机器学习方法是利用机器学习模型基于数据学习直线模式的模型表达，如文献[19]利用随机森林识别线性模式，文献[20]利用图卷积神经网络识别建筑物群模式。其中，面向复杂多样的建筑物模式，实际应用中需要依据用户要求

定义不同的直线模式。相比模板匹配或机器学习方法，邻近图方法因其灵活定义的结构化规则，能较好适应用户的个性化需要，是当前研究中普遍采用的方法。但是，低效的图遍历策略制约了邻近图方法面向大规模区域的应用。

知识图谱利用图模型表达实体和实体间关系，发展有高效的知识图谱存储和推理工具，并已在地理知识抽取、融合和推理等方面应用广泛[21-24]。图数据库作为知识图谱存储和推理的重要工具，它将实体间关系作为实体的关系列表存储，图谱推理时无需消耗计算资源建立实体与实体间关系的匹配，能实现与图谱规模无关的高效推理[25]。因此，若将建筑物模式识别的邻近图数据模型转换为知识图谱表示的图数据模型，构建面向建筑物模式识别的知识图谱，利用图数据库实现知识图谱的存储和推理，就能快速挖掘建筑物群中的直线模式。因此，本研究试图：(1)建立知识图谱表示的图数据模型与邻近图数据模型的联系；(2)构建面向直线型建筑物模式识别的知识图谱；(3)将直线型模式识别的结构化规则转化为适宜知识图谱推理的规则，基于知识图谱进行规则推理高效识别直线型建筑物模式。

## 1. 知识图谱表示的属性图数据模型

属性图通过包含属性的顶点和边表达图数据，符合人类认知习惯，是当前知识图谱表示采用最广泛的数据模型之一[21]。属性图记为 $G=(V,E)$，$V$ 为顶点集合 $V=\{v_1,v_2,...,v_m\}$，$v_m$ 表示实体；$E$ 为边集合 $E=\{e(v_i,v_j),v_i\in V,v_j\in V\}$，$e(v_i, v_j)$ 表示实体 $v_i$ 和 $v_j$ 间关系。实体 $v_m$ 包含标签和属性，标签表示实体所属概念；关系 $e(v_i, v_j)$ 包含类型和属性。属性图数据模型规定：实体 $v_m$ 可以有多个标签，关系 $e(v_i, v_j)$ 只能有一个类型。基于上述定义，可利用属性图数据模型表达图 1 中建筑物的面积(*Area*)，建筑物间邻近关系(*Has_Proxi*)、面积相似关系(*Area_Sim*)和面积相似程度的(*SimDeg*)，如图 1 右侧。对比基于邻近图数据模型的表示(图 1 左侧)可知，邻近图数据模型中边仅表示建筑物间存在邻近关系，建筑物间其它类型关系和关系属性均作为边的属性表达；属性图数据模型中边则可以表示建筑物间的各类关系，如邻近、面积相似等。需要说明的是：实际应用中邻近图数据模型和属性图数据模型依据需要可以有不同的表达形式。

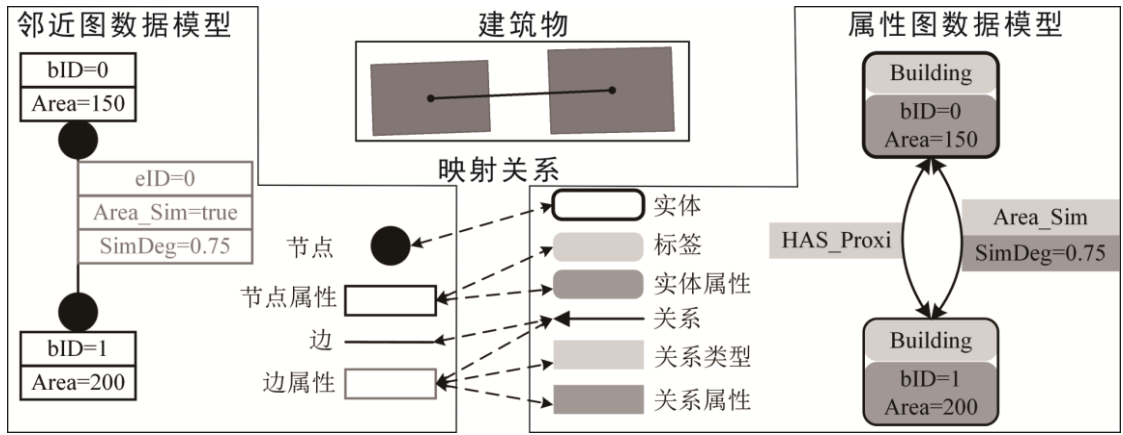

图 1. 知识图谱表示的属性图数据模型

Fig 1. Property graph model for knowledge graph

## 2. 面向直线型建筑物模式识别的知识图谱构建

直线型建筑物模式可认为是一组满足格式塔原则(邻近、相似)的沿直线排列建筑物[8, 11, 18]。因此，构建面向直线型建筑物模式识别的知识图谱需要：(1)定义建筑物间"邻近"、"相似"和"沿直线排列"的空间关系；(2) 利用知识图谱表达建筑物间"邻近"、"相似"和"沿直线排列"的空间关系。

### 2.1 直线排列空间关系定义

(1)邻近关系

RNG 能用更少边较完整地体现建筑物群的直线排列关系[18]，故利用 RNG 表达建筑物间邻近关系。另外，建筑物分布通常与所处地理环境有关，本文构建 RNG 时考虑了邻近道路为约束条件(图 2)。RNG 记为 $RNG=(RV,RE)$，$RV$ 为顶点集合 $RV=\{rv_1, rv_2, ..., rv_m\}$，$rv_m$ 表示建筑物 $B_m$；$RE$ 为边集合 $RE=\{re(rv_i, rv_j), rv_i \in RV, rv_j \in RV\}$，$re(rv_i, rv_j)$ 表示建筑物 $B_i$ 和 $B_j$ 邻近，$re(v_i, v_j)$长度($Le$)为建筑物 $B_i$ 和 $B_j$ 间最短距离。

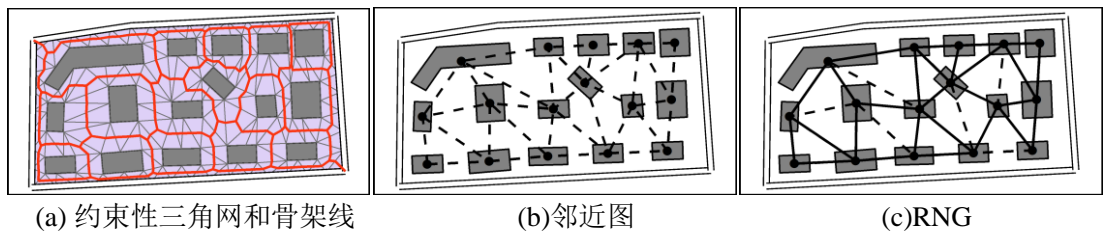

(a) 约束性三角网和骨架线　　(b)邻近图　　(c)RNG

图 2. 建筑物邻近关系

Fig 2. Proximity relations between buildings

(2) 相似关系

建筑物 $B_i$ 和 $B_j$ 相似记为 $S(B_i, B_j)$，可用建筑物间图形特征(如大小、方向和形状等)的相似性表示，定义见表1和公式(1)，$\delta_1$、$\delta_2$ 和 $\delta_3$ 为阈值[26]。

$$S(B_i, B_j) =_{def} \{A_r \leq \delta_1, O_r \leq \delta_2, E_r \leq \delta_3\} \tag{1}$$

表 1. 相似关系定义
Tab 1. Definition of similarity

| 参数 | 描述 | 计算方法 |
|---|---|---|
| $A_r$ | 建筑物 $B_i$ 和 $B_j$ 面积差异 | $A_r = \max(A_i, A_j)/\min(A_i, A_j)$，$A_i$ 和 $A_j$ 为 $B_i$ 和 $B_j$ 面积 |
| $O_r$ | 建筑物 $B_i$ 和 $B_j$ 最小绑定矩形 $SBR_i$ 和 $SBR_j$ 方向差异 | $O_r = \begin{cases} \|O_i - O_j\| & (\|O_i - O_j\| < 90) \\ 180 - \|O_i - O_j\| & (\|O_i - O_j\| \geq 90) \end{cases}$，$O_i$ 和 $O_j$ 为 $SBR_i$ 和 $SBR_j$ 主轴与水平方向夹角 |
| $E_r$ | 建筑物($B_i$ 和 $B_j$)边数差异 | $E_r = \max(E_i, E_j)/\min(E_i, E_j)$，$E_i$ 和 $E_j$ 为 $B_i$ 和 $B_j$ 边数 |

(3) 沿直线排列空间关系

三元组是直线型模式的最小单元，直线模式可以看成是一个或多个三元组直线模式构成。因此，首先定义三个建筑物沿直线排列的空间关系，可用 RNG 两邻接边的方向差异、距离差异和建筑物间正对面积比表示，定义见表 2[12, 18]。考虑到城区中建筑物间可能距离较近，即 RNG 边长度 $Le$ 可能较小，当两邻接的 RNG 边长度均较小时，可一定程度认为两 RNG 边长度差异($D_L$)较小。因此，若 $Le \leq T_d$，则 $Le = T_d$；$T_d$ 为阈值，可依据地图上图形之间可辨识的最小距离阈值确定(一般设为 0.2 mm)[27]。给定 RNG 邻接边关联的建筑物三元组 $\{B_i, B_j, B_k\}$，$\{B_i, B_j, B_k\}$ 沿直线排列的定义见公式(2)，记为 $Str(B_i, B_j, B_k)$，$\eta_1$、$\eta_2$ 和 $\eta_3$ 为阈值。三个以上建筑物沿直线排列的空间关系则是以三元组为基础，结合直线模式识别过程通过依次添加建筑物判断，定义见章节 3.1。

$$Str(B_i, B_j, B_k) =_{def} \{D_o \leq \eta_1, D_L \leq \eta_2, FR_{ij} \geq \eta_3, FR_{jk} \geq \eta_3\} \tag{2}$$

表 2. 三个建筑物沿直线排列空间关系定义
Tab 2. Definition of linear arrangement

| 参数 | 描述 | 计算方法 |
|---|---|---|
| $D_o$ | RNG 两邻接边 $re(v_i, v_j)$ 和 $re(v_j, v_k)$ 的方向差异 | $e(v_i, v_j)$ 和 $e(v_j, v_k)$ 夹角 |
| $D_L$ | RNG 两邻接边 $re(v_i, v_j)$ 和 $re(v_j, v_k)$ 的长度差异 | $D_L = \max(Le_{ij}, Le_{jk})/\min(Le_{ij}, Le_{jk})$，$Le_{ij}$ 和 $Le_{jk}$ 为 $re(v_i, v_j)$ 和 $re(v_j, v_k)$ 长度； |
| $FR_{ij}$ | 建筑物 $B_i$ 和 $B_j$ 最小绑定矩形 $SBR_i$ 和 $SBR_j$ 的正对面积比 | 以 $SBR_i$ 和 $SBR_j$ 之一为参考，分别在其主方向和法方向做 $SBR_i$ 和 $SBR_j$ 投影，求投影最大重叠长度与该方向上总投影长度之比 |

## 2.2 知识图谱构建

本研究面向直线型建筑物模式识别需要，采用自底而上的方式按需扩展构建知识图谱[21]，记为 $G=(V, E)$。$V$ 为顶点集合 $V=\{v_1, v_2, ..., v_m\}$，$v_m$ 表示实体，$v_m$ 存在属性 $ID$，$ID$ 为实体唯一标识；$E$ 为边集合 $E=\{e(v_i, v_j), v_i \in V, v_j \in V\}$，$e(v_i, v_j)$ 表示实体 $v_i$ 和 $v_j$ 间关系。基于章节 2.1 定义，知识图谱表达的实体是建筑物，故实体 $v_m$ 存在标签 *Building*，表示 $v_m$ 是建筑物实体。另外，需表达建筑物间"邻近"、"相似"和"沿直线排列"的空间关系。

(1) 邻近关系表达

两建筑物 $B_i$ 和 $B_j$ 邻近可由实体 $v_i$ 和 $v_j$ 存在邻近关系(*HAS_Proxi*)表达。若两 $B_i$ 和 $B_j$ 邻近，则向知识图谱 $G$ 的边集合 $E$ 中添加边 $e(v_i, v_j)$，$e(v_i, v_j)$ 存在关系类型 *HAS_Proxi*，表示 $B_i$ 和 $B_j$ 邻近；存在属性 *Proxi_Ori*，表示 $B_i$ 和 $B_j$ 邻接边的方向。

(2) 相似关系表达

两建筑物 $B_i$ 和 $B_j$ 相似可由实体 $v_i$ 和 $v_j$ 存在相似关系(*HAS_Sim*)表达。其中，邻近且相似的建筑物才可能构成直线模式，故本研究仅考虑邻近两建筑物间的相似性。若 $B_i$ 和 $B_j$ 邻近且相似，则向知识图谱 $G$ 的边集合 $E$ 中添加边 $e(v_i, v_j)$，$e(v_i, v_j)$ 存在关系类型 *HAS_Sim*，表示 $B_i$ 和 $B_j$ 相似。

(3) 沿直线排列的空间关系表达

沿直线排列表达了邻近三个建筑物间的空间关系，无法直接利用两个建筑物实体间关系表达，本研究利用建筑物实体属性 *pIDList* 表示，*pIDList* 表示建筑物实体所属沿直线排列三元组的编号列表。过程如下：首先，识别建筑物群中所有沿直线排列的建筑物三元组，每个三元组存在唯一标识 *pID*；其次，记录每一个建筑物实体所属的直线排列三元组 *pID*，添加至属性 *pIDList* 中。例如，图 3 存在 3 个沿直线排列的建筑物三元组，*pID* 分别为 0，1 和 2，建筑物 $B_3$ 属于三元组 1、2，故 $B_3$ 对应的实体 $v_3$ 的 *pIDList* 属性值为[1, 2]。若三个建筑物实体 *pIDList* 属性值存在相同元素，表示三个建筑物实体存在沿直线排列的空间关系，如图 3 中建筑物 $B_3$、$B_4$ 和 $B_5$。

基于上述分析，面向直线型建筑物模式识别的知识图谱表达见表3和图3。

表 3. 面向直线型建筑物模式识别的知识图谱表达要素
Tab 3. The elements of knowledge graph for linear building pattern recognition

| 分类 | 要素 | 描述 |
| --- | --- | --- |
| 实体 | $v_m$ | 表示建筑物实体 $B_m$ |
| 标签 | Building | 表示实体 $v_m$ 是一个建筑物 |
| 实体属性 | ID | 实体唯一标识 |
|  | pIDList | 标签为 Building 的实体属性，表示建筑物所处的沿直线排列三元组的 pID 列表 |
| 关系 | $e(v_i, v_j)$ | 表示两个建筑物实体 $v_i$ 和 $v_j$ 存在 HAS_Proxi 或 HAS_Sim 关系 |
| 关系类型 | HAS_Proxi | 表示两个建筑物实体间关系类型为邻近 |
|  | HAS_Sim | 表示两个建筑物实体间关系类型为相似 |
| 关系属性 | EOri | 关系 Has_Proxi 属性，表示两个建筑物邻接边的方向 |

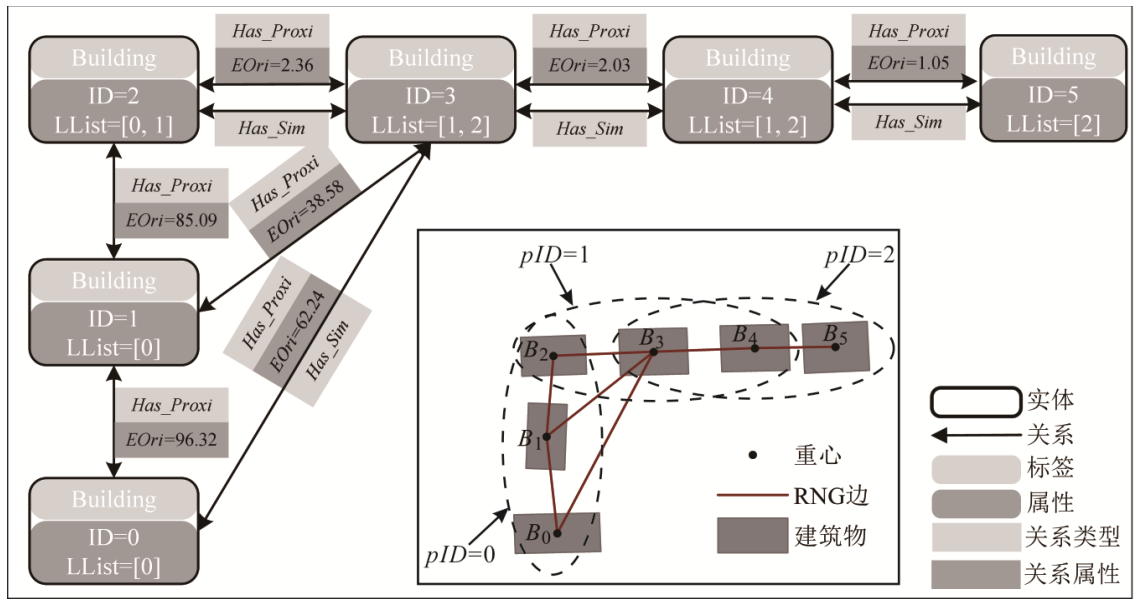

图 3. 面向直线型建筑物模式识别的知识图谱表达示意图

Fig 3. An example knowledge graph for linear building pattern recognition

# 3. 基于规则推理的直线型建筑物模式识别
## 3.1 直线型建筑物模式识别的规则

直线型建筑物模式可认为是一组满足格式塔原则(邻近、相似)的沿直线排列建筑物，三元组是直线型建筑物模式的最小单元[8, 17]。给定建筑物三元组$\{B_i, B_j, B_k\}$，$\{B_i, B_j, B_k\}$构成三元组直线模式 $LP\{B_i, B_j, B_k\}$ 的定义见公式(3)。

$$LP(B_i, B_j, B_k) =_{def} \{P(B_i, B_j) \wedge P(B_j, B_k) \wedge S(B_i, B_j) \wedge S(B_j, B_k) \wedge Str(B_i, B_j, B_k)\} \quad (3)$$

一个直线模式可认为是一个或多个包含两个重复元素的三元组直线模式构成，因此，依据一定约束条件合并识别的三元组直线模式即为最终的直线模式识别结果。其中，合并两个三元组直线模式应保证合并后直线模式延伸方向变化小于阈值 (Align_Rule)[11]。若给定包含两个重复元素的三元组直线模式 $LP\{B_i, B_j, B_k\}$ 和 $LP\{B_j, B_k, B_p\}$，关联 $B_i, B_j, B_k$ 和 $B_p$ 的邻近边分别为 $re(v_i, v_j)$，$re(v_j, v_k)$ 和 $re(v_k, v_p)$，参考文献[11]，Align_Rule 需保证 $re(v_i, v_j)$ 和 $re(v_k, v_p)$ 角度差异小于阈值，即 $D_o(re(v_i, v_j), re(v_k, v_p)) \leq \eta_1$，$D_o$ 为两条边方向差异，定义见表 3，$\eta_1$ 为阈值条件。

因此，基于规则识别直线模式的过程如下：(1) 识别建筑物群中所有三元组直线模式；(2) 判断不同三元组直线模式是否可合并，将可合并的三元组直线模式归入同一列表，三元组直线模式可合并的条件定义如下：存在 2 个相同元素且满足 Align_Rule；(3)合并同一列表中三元组直线模式为最终模式识别结果。

## 3.2 基于 Cypher 查询语言的直线型模式识别推理规则表达

基于知识图谱进行规则推理需将模式识别的结构化规则表达为适宜知识图谱推理的规则。Neo4j 是一种广泛采用的知识图谱存储与管理数据库；同时，Neo4j 提供 Cypher 查询语言，支持用户自定义规则进行高效查询[25]。因此，本研究利用 Neo4j 存储知识图谱，利用 Cypher 查询语言表达直线模式推理的规则。依据章节 3.1 定义，基于 Cypher 查询语言的直线型模式推理规则应包括三部分：(1)基于公式(3)定义识别知识图谱中三元组直线模式，并在知识图谱中创建标签为 *Triple_Pattern* 的实体表示三元组直线模式实体，实体包含属性 *bIDList* 和 *OriList*，记录三元组直线模式包含的建筑物 ID 列表和两邻接边的角度(*EOri*)；(2)判断知识图谱中不同三元组直线模式是否可合并，若两个三元组直线模式满足合并条件，为两个三元组直线模式实体添加 *Align_true* 关系；(3)合并所有存在 *Align_true* 关系的三元组直线模式为最终直线模式识别结果。其中，三元组存在相同元素的数量可通过 apoc.coll.intersectio(stLP1: Triple_Pattern, stLP2: Triple_Pattern)获取。详细直线模式识别推理规则如下，在 Neo4j 图数据库中进行规则推理获取的直线排列结果见图 4。

**直线型模式识别推理规则：**

MATCH ($B_1$:Building)-[rp1:HAS_Proxi]->($B_2$:Building)-[rp2:HAS_Proxi]->($B_3$: Building),
　　　($B_1$: Building)-[rs1:HAS_Sim]->($B_2$: Building)-[rs2:HAS_Sim]->($B_3$: Building)

WHERE size (apoc.coll.intersection (apoc.coll.intersection ($B_1$.pIDList, $B_2$.pIDList), $B_3$.pIDList)) > 0

    WITH $B_1$, $B_2$, $B_3$, rp1, rp2

MERGE (tLP:Triple_Pattern {bIDList: ([$B_1$.ID, $B_2$.ID, $B_3$.ID]), OriList: [rp1.EOri, rp2.EOri]})

MATCH (stLP1: Triple_Pattern), (stLP2: Triple_Pattern)

WHERE stLP1<>stLP2 AND size(apoc.coll.intersection (stLP1.bIDList, stLP2.bIDList))>=2
AND (abs(stLP2.OriList[1]-stLP1.OriList[0])<= $\eta_1$ OR (180-abs(stLP2.OriList[1]-stLP1.OriList[0]))<= $\eta_1$ ) AND (abs(stLP2.OriList[1]-stLP1.OriList[1])<= $\eta_1$ OR (180-abs(stLP2.OriList[1]- stLP1.OriList[1]))<= $\eta_1$ ) AND (abs(stLP2.OriList[0]-stLP1.OriList[0])<= $\eta_1$ OR (180-abs(stLP2.OriList[0]- stLP1.OriList[0]))<= $\eta_1$ ) AND (abs(stLP2.OriList[0]-stLP1.OriList[1])<= $\eta_1$ OR (180-abs(stLP2.OriList[0]- stLP1.OriList[1]))<= $\eta_1$ )

CREATE (stLP1)-[r:Align_true]->(stLP2)

MATCH resLP=(rLP1: Triple_Pattern)-[: Extend_true *0..]-(rLP2: Triple_Pattern)

RETURN resLP

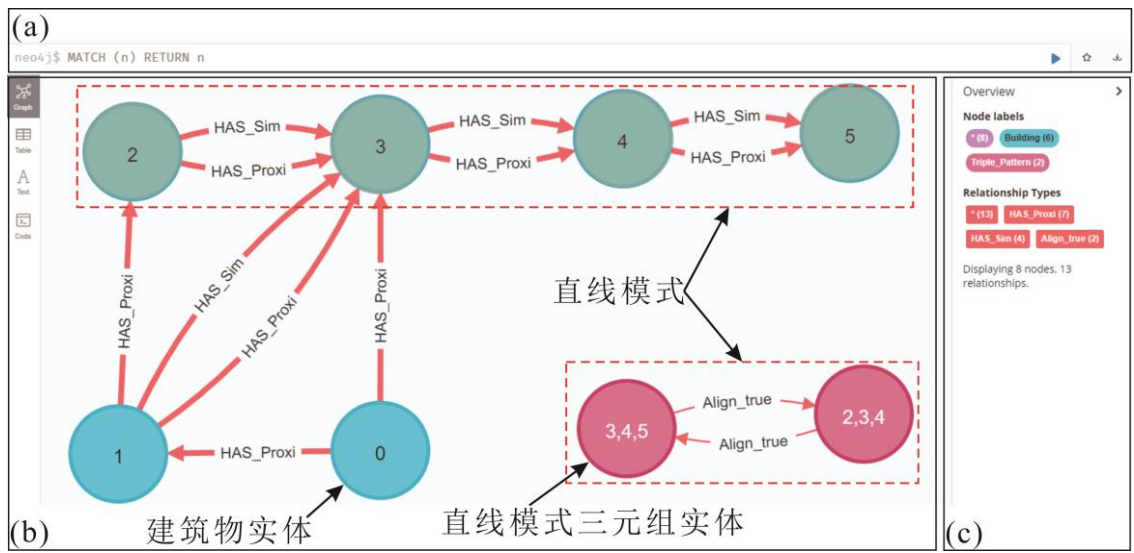

图 4. 在 Neo4j 中利用 Cypher 查询语言进行规则推理识别直线型建筑物模式

Fig 4. Linear building pattern recognition via Cypher in Neo4j

## 4. 实验

### 4.1 实验数据

选取 OpenStreetMap(OSM)中 Dresden 部分地区建筑物作为实验数据(图 5)，建筑物的数量($B\_Count$)、平均面积($Ave\_A$)、平均边数($Ave\_E$)和边数不大于 8 的建筑物占比($Rate(E \leq 8)$)统计结果见表 4。由表 4 可知：(1) Dresden 地区建筑

物 *Ave_E* 为 5.41，*Rate*(*E* ≤ 8) 比例为 90.47%，说明 Dresden 地区大部分建筑物形状简单，可以用边数度量建筑物间形状差异；(2)实验区域内建筑物和 Dresden 地区建筑物在指标 *Ave_A*、*Ave_E* 和 *Rate*(*E* ≤ 8) 上均相近，说明选取的实验区域具有一定代表性。

表 4. 建筑物形态特征统计分析
Tab 4. The statistical results for characteristics of buildings

|  | *B_Count* | *Ave_A*/m² | *Ave_E* | *Rate*(*E* ≤ 8) |
|---|---|---|---|---|
| Dresden 地区 | 158775 | 345.224 | 5.41 | 90.47% |
| 实验区域 | 1286 | 416.60 | 5.36 | 92.53% |

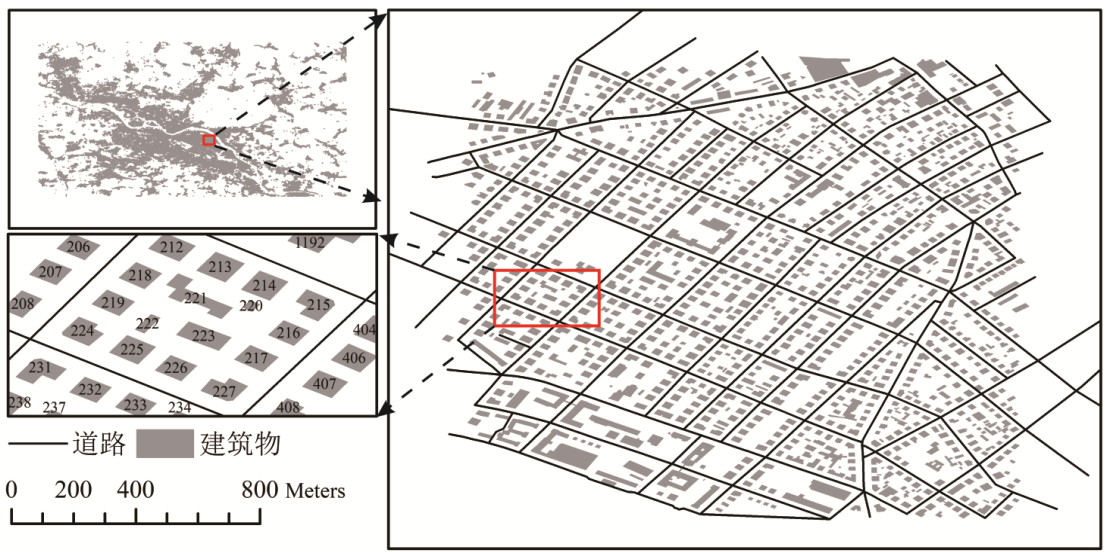

图 5. 实验数据

Fig 5. Experiment data

## 4.2 知识图谱构建结果

依据章节 2 定义，构建知识图谱表达建筑物间邻近、相似且沿直线排列的空间关系。其中，邻近关系用 RNG 表示；建筑物间相似则用建筑物间大小、方向和形状的相似性小于设定的阈值($\delta_1$，$\delta_2$ 和 $\delta_3$)表示，定义见表 1 和公式 1，参考文献[26]，设置 $\delta_1=2$，$\delta_2=20$，$\delta_3=1.5$；沿直线排列的空间关系则用 RNG 两邻接边的方向差异、距离差异和建筑物间正对面积比小于设定的阈值($\eta_1$，$\eta_2$ 和 $\eta_3$)表示，定义见表 2 和公式 2，参考文献[12, 18]，设置 $\eta_1=15$，$\eta_2=2$，$\eta_3=0.3$。知识图谱构建结果见图 6：图 6(a)中不同建筑物实体依据定义的邻近关系形成聚群；图 6(b)是构建的知识图谱局部放大图，表达了图 5(c)所示局部区域内建筑物间邻近、相似和沿直线排列的空间关系。

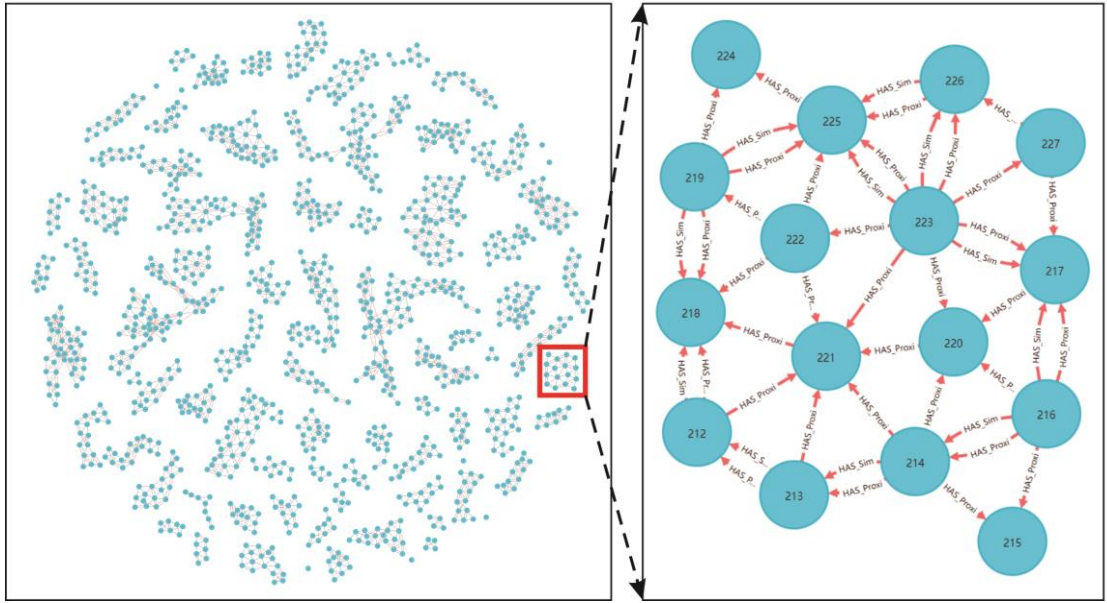

图 6. 面向直线型建筑物模式识别的知识图谱

Fig 6. The built knowledge graph for linear build pattern recognition

## 4.3 直线模式识别结果

(1) 识别效果评价

为验证本文方法识别效果，基于定义的推理规则在构建的知识图谱中进行推理识别实验区域内直线模式，结果见图 7。依据认知心理学实验要求，将本文方法识别结果与人眼视觉认知结果对比；人眼视觉认知结果是由 5 名从事地图学研究的硕(博)士生利用视觉感知和投票后获取。其中，与人视觉认知一致的直线模式为 127 个，与人视觉认知不一致的直线模式为 5 个，漏识别的直线模式为 12 个，识别召回率(*recall*)为 127/(127+5)=91.36%，识别准确率(*precision*)为 127/(127+12)=96.21%。上述结果表明本文基于知识图谱能有效识别实验区域内大部分直线模式，识别结果符合视觉认知。

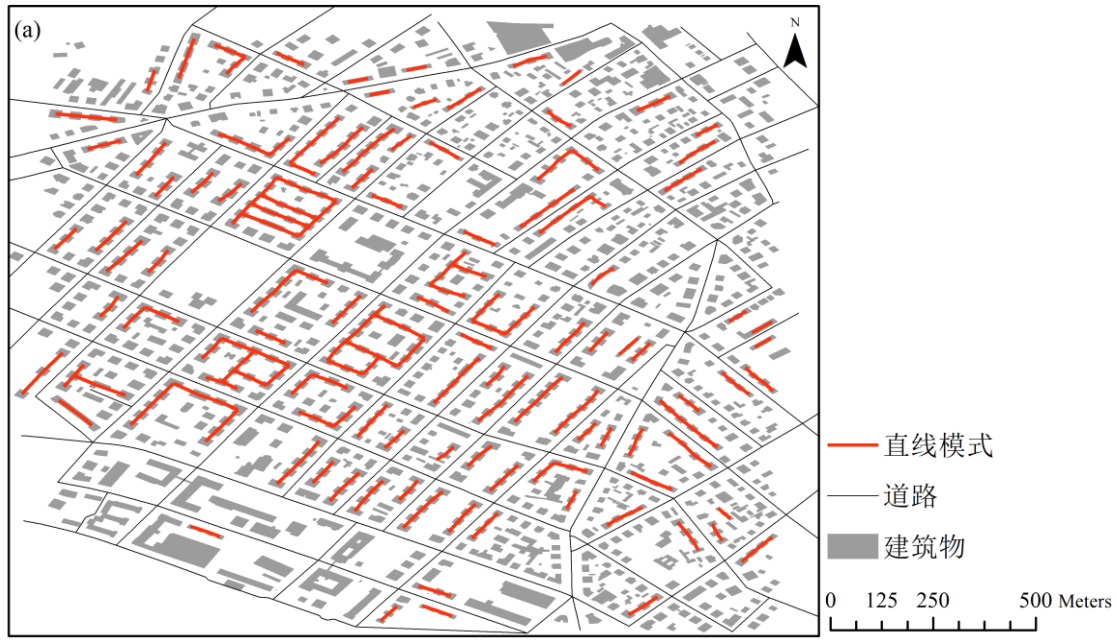

图 7. 直线模式识别结果
Fig 7. The results for linear building pattern recognition

(2) 识别效率评价

基于知识图谱识别直线模式的主要目的是提高模式识别的效率。因此，利用邻近图数据模型预先存储章节 2 定义的建筑物间邻近、相似和沿直线排列空间关系，利用图同构方法基于预先构建的邻近图数据模型识别直线模式，对比本研究基于构建的知识图谱进行推理识别直线模式的耗时($t$)。实验环境：图同构方法在 AE10.2 基础上利用 C#二次开发实现，基于知识图谱的规则推理则基于 Neo4j 1.4.15 实现，实验平台是一台 CPU 为 Intel(R) Core(TM) i5-8265U CPU @1.60GHz、内存为 8GB、操作系统为 Windows 10(64 位)的计算机。两种方法均独立进行 10 组实验，耗时最小值($Min\_t$)、最大值($Max\_t$)、平均值($Ave\_t$)和标准差($Std\_t$)的统计结果见表 5。由表 5 可知，本文方法平均耗时为 0.642s，小于图同构算法的平均耗时 3.839s，算法效率平均提高了 5.98 倍。上述结果表明，本文方法相比已有方法能更高效的识别直线型建筑物模式。

表 5. 效率统计分析
Tab 5. Statistical analysis of efficiency

|  | $Min\_t$/s | $Max\_t$/s | $Ave\_t$/s | $Std\_t$/s |
| --- | --- | --- | --- | --- |
| 本文方法 | 0.454 | 0.925 | 0.642 | 0.159 |
| 图同构方法 | 3.271 | 4.282 | 3.839 | 0.383 |

# 5. 讨论
## 5.1 知识图谱构建策略对识别结果的影响分析

本研究利用知识图谱扩展性强的特点采用自底向上的方式构建知识图谱，该方式的好处是无需经过顶层设计，可依据应用灵活构建和扩展知识图谱。但是，构建知识图谱不同，其推理规则和模式识别效率也会不同。本研究中建筑物间相似、沿直线排列的空间关系均提前计算后作为实体间关系或属性表达，实际上这些关系亦可基于实体间属性进行规则推理获取。例如，基于公式(1)和(2)定义，可构建表达要素如表 6 所示的知识图谱，利用下文所示推理规则可获取实体间相似和沿直线排列关系，在此基础上结合前文直线模式识别推理规则亦可识别直线模式。另外，利用邻近图数据模型预先存储表 6 中定义的建筑物属性和邻近关系，利用图同构方法基于预先构建的邻近图数据模型识别直线模式，分别统计两种方法耗时，结果见表 7。由表 7 可知，若基于表 6 构建的知识图谱识别直线模式，本文方法平均耗时为 1.109s，小于图同构方法的平均耗时 28.871s，效率提高 26.03 倍。另外，对比表 5 和表 7 可知，构建的知识图谱不同，模式识别效率会不同；若将知识图谱构建阶段涉及的部分实体间关系在推理阶段利用规则推理获取，尽管会增加规则推理耗时，但也会有效减少知识图谱构建阶段相应实体间关系计算的耗时，即可进一步提高模式识别效率，如由 5.98 倍提高至 26.03 倍。另外，若实体间关系在推理阶段利用规则推理获取，一定程度可增加用户使用的灵活性，用户能通过修改推理规则涉及的参数获取不同条件定义的直线模式。实际应用中，应依据需要构建知识图谱。

表 6. 面向直线型建筑物模式识别的知识图谱表达要素
Tab 6. The elements of knowledge graph for linear building pattern recognition

| 分类 | 要素 | 描述 |
| --- | --- | --- |
| 实体 | $v_m$ | 表示建筑物实体 $B_m$ |
| 标签 | *Building* | 表示实体 $v_m$ 是一个建筑物 |
| 实体属性 | *ID* | 实体唯一标识 |
| | *Area* | 建筑物实体属性，表示建筑物面积 |
| | *BOri* | 建筑物实体属性，表示建筑物最小绑定矩形主轴与水平方向夹角 |
| | *EdgeCount* | 建筑物实体属性，表示建筑物边数 |
| 关系 | $e(v_i, v_j)$ | 表示两个建筑物实体 $v_i$ 和 $v_j$ 存在 *HAS_Proxi* 或 *HAS_Sim* 关系 |
| 关系类型 | *HAS_Proxi* | 表示两个建筑物实体间关系类型为邻近 |

| 关系属性 | EOri | 关系 *Has_Proxi* 属性，表示两个建筑物邻接边的方向 |
| --- | --- | --- |
| | Length | 关系 *Has_Proxi* 属性，表示邻接的两个建筑物间最短距离 |
| | FR | 关系 *Has_Proxi* 属性，表示邻接的两个建筑物的正对面积 |

**建筑物实体间相似和沿直线排列关系获取的推理规则：**

MATCH ($B_1$:Building)-[rp1:HAS_Proxi]->($B_2$:Building)

WHERE ((B1.Area/B2.Area<= $\delta_1$ AND B1.Area/B2.Area>=1) OR (B2.Area/B1.Area<= $\delta_1$ AND B2.Area/B1.Area>=1)) AND (abs(B1.BOri-B2.BOri)<= $\delta_2$ OR (180 - abs(B1.BOri-B2.BOri))<= $\delta_2$ ) AND ((B1.EdgeCount/B2.EdgeCount<= $\delta_3$ AND B1.EdgeCount/B2.EdgeCount>=1) OR (B2.EdgeCount/B1.EdgeCount<= $\delta_3$ AND B2.EdgeCount/B1.EdgeCount>=1))

CREATE (B1)-[r:HAS_Sim]->(B2)

MATCH (B1:Building)-[rp1:HAS_Proxi]->(B2:Building)-[rp2:HAS_Proxi]->(B3: Building)

WHERE (abs(rp1.EOri-rp2.EOri)<= $\eta_1$ OR (180 - abs(rp1.EOri-rp2.EOri))<= $\eta_1$ ) AND ((rp1.Length/rp2.Length<= $\eta_2$ AND rp1.Length/rp2.Length>=1) OR (rp2.Length/rp1.Length<= $\eta_2$ AND rp2.Length/rp1.Length>=1)) AND rp1.FR>= $\eta_3$ AND rp2.FR>= $\eta_3$

Return B1, B2, B3

表 7. 效率统计分析

Tab 7. Statistical analysis of efficiency

|  | *Min_t*/s | *Max_t*/s | *Ave_t*/s | *Std_t*/s |
| --- | --- | --- | --- | --- |
| 本文方法 | 0.871 | 1.466 | 1.109 | 0.194 |
| 图同构方法 | 26.718 | 30.937 | 28.871 | 1.373 |

## 5.2 数据规模对模式识别效率的影响分析

选取 5 个不同规模的数据集，分别利用本文方法基于构建的知识图谱和利用图同构算法基于预先构建的邻近图数据模型提取直线模式，分析数据规模对模式识别效率的影响。其中，统计知识图谱中实体数量(*vCount*)、关系数量(*eCount*)表示数据规模，并统计两个方法的平均耗时(*Ave_t*)和平均耗时比率(*eRate*)表示方法效率，结果见表 8 和图 8。由表 8 和图 8 可知：(1) 基于图同构方法识别直线模式的平均耗时会随着实体数量和关系数量的增长呈现近似指数的增长，而本文方法平均耗时随实体数量和关系数量的增长呈现近似线性的增长，这说明本文方法相比图同构方法算法耗时不会因为数据规模的增长而显著增加。(2) 针对小规模数据集，图同构算法平均耗时较本文方法小，如针对数据

集 Dataset_1(实体数为 36，关系数为 114)，本文方法平均耗时为 0.149s，而图同构方法平均耗时为 0.029s，小于本文方法；而对比图 8(a)和图 8(c)可知，当实体数量超过约 600 时，本文方法平均耗时会小于图同构方法，且随着数据规模的增大，这种差异会逐渐增大，如对于数据集 Dataset_4(实体数 1295)，本文方法较图同构方法效率提高 5.98 倍，对于数据集 Dataset_5(实体数为 3566)，本文方法较图同构方法效率提高 31.49 倍。另外，当数据集实体数为 3566，关系数为 7538 时，本文方法模式识别的平均耗时为 0.946s，小于 1s。以上均说明本文方法面向大规模区域的建筑物模式识别更高效，而面向小规模区域的建筑物模式分析时图同构算法则更高效。(3) 由图 8(c)和 8(d)可知，本文方法平均耗时随实体数量和关系数量的增长变化趋势相近，这说明实体数量和关系数量的增加均会对算法的耗时产生影响，且影响相近。

表 8. 数据规模对效率的影响分析
Tab 8. Analysis of the influence of data size on efficiency

|  | *vCount* | *eCount* | *Ave_t/s* | | *eRate* |
|---|---|---|---|---|---|
|  |  |  | 本文方法 | 图同构方法 |  |
| Dataset_1 | 36 | 114 | 0.149 | 0.029 | 0.19 |
| Dataset_2 | 241 | 619 | 0.321 | 0.181 | 0.56 |
| Dataset_3 | 685 | 1526 | 0.411 | 0.465 | 1.13 |
| Dataset_4 | 1295 | 3266 | 0.642 | 3.8389 | 5.98 |
| Dataset_5 | 3566 | 7538 | 0.946 | 29.792 | 31.49 |

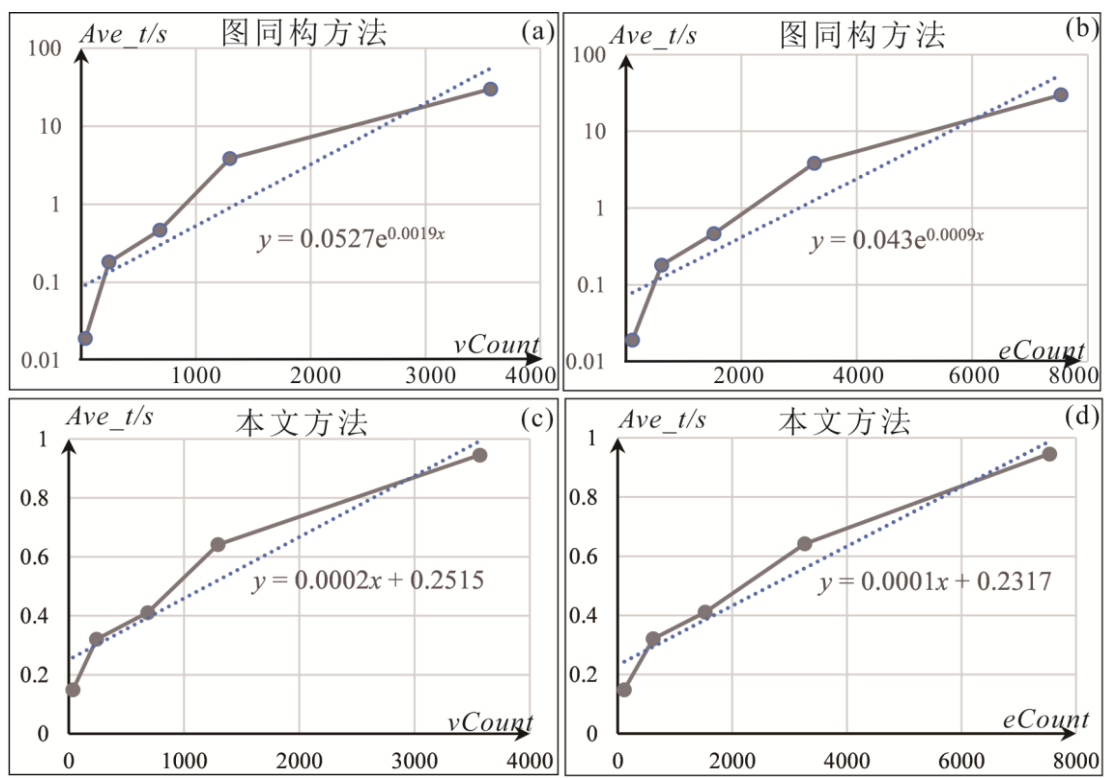

图 8. 数据规模对效率影响分析
Fig 8. Efficiency analysis of the impact of data size

# 6. 结论

  为提高直线型建筑物模式识别的效率，本文利用知识图谱表达建筑物间邻近、相似和沿直线排列的空间关系，利用规则推理识别建筑物群中直线模式。实验结果证明，本文方法能更高效地识别建筑物群中直线模式，识别结果符合视觉认知。本文方法为后续利用知识图谱进行地图空间表达和推理提供了有效的理论支撑。但是，视觉认知模式具有较大的灵活性，本文基于规则识别直线模式会导致部分模式的漏识别和误识别；同时，基于规则识别模式只能识别用户预先定义的模式，无法获取建筑物群中潜在的模式特征。利用知识图谱进行归纳推理识别和理解建筑物群潜在的模式特征是未来研究的重点。

# 参考文献

[1] Carmona M, Heath T & Oc T, et al. Public Places, Urban spaces: The Dimensions of Urban Design[M]. 2003, Oxford: Architectural Press.
[2] Mao B, Harrie L & Ban Y. Detection and typification of linear structures for dynamic visualization of 3D city models[J]. Computers, Environment and Urban Systems, 2007, 6(3): 233-244.

[3] Zhang L, Deng H & Chen D. A spatial cognition-based urban building clustering approach and its applications. International Journal of Geographical Information Science, 2013, 27(4), 721–740.

[4] Ruas A & Holzapfel F. Automatic characterization of building alignments by means of expert knowledge[C]//21th ICC conference. 2003, 1604-1616.

[5] Zhang X, Stoter J & Ai T, et al. Automated evaluation of building alignments in generalized maps. International Journal of Geographical Information Science, 2013, 27(8): 1550–1571.

[6] Renard J & Duchêne C. Urban structure generalization in multi-agent process by use of reactional agents. Transactions in GIS, 2014, 18(2): 201–218.

[7] Rainsford D & Mackaness W. Template matching in support of generalization of rural buildings [C]. The 10th International Symposium on Spatial Data Handling, Springer, Berlin, 2002, 137–151.

[8] Du S, Shu M & Feng C C. Representation and discovery of building patterns: a three-level relational approach[J]. International Journal of Geographical Information Science, 2015, 30(6): 1161-1186.

[9] Du S, Luo L & Cao K, et al. Extracting building patterns with multilevel graph partition and building grouping[J]. ISPRS Journal of Photogrammetry and Remote Sensing, 2016, 122: 81-96.

[10] 行瑞星, 武芳, 巩现勇等. 建筑群组合直线模式识别的模板匹配方法[J]. 测绘学报, 2021, 50(6): 800-811. (Xing R, Wu F & Gong X, et al. The template matching approach to combined collinear pattern recognition in building groups[J]. Acta Geodaetica et Cartographica Sinica, 2021, 50(6): 800-811.)

[11] Zhang X, Ai T & Stoter J. Building pattern recognition in topographic data: examples on collinear and curvilinear alignments[J]. Geoinformatica, 2013, 17(1):1-33.

[12] Wang X & Burghardt D. A typification method for linear building groups based on stroke simplification[J]. Geocarto International, 2021, 36(15): 1732-1751.

[13] 巩现勇, 武芳, 钱海忠等. 建筑群多连通直线模式的参数识别方法[J]. 武汉大学学报·信息科学版, 2014, 39(3): 335-339. (Gong X, Wu F & Qian H, et al. The parameter discrimination approach to multi-connected linear pattern recognition in building groups[J]. Geomatics and Information Science of Wuhan University, 2014, 39(3): 335-339.)

[14] 巩现勇, 武芳. 城市建筑群网格模式的图论识别方法[J]. 测绘学报, 2014, 43(9): 960-968. (Gong X, Wu F. The graph theory approach to grid pattern recognition in urban building groups[J]. Acta Geodaetica et Cartographica Sinica, 2014, 43(9): 960-968.)

[15] Wei Z, Ding S & Cheng L, et al. Linear building pattern recognition in topographical maps combining convex polygon decomposition[J]. Geocarto International, 2022: 1-25.

[16] 魏智威, 丁愫, 童莹等. 格式塔原则与图形凸分解结合的建筑物群直线模式识别方法[J]. 2023, 52(1): 117-128. (Wei Z, Ding S & Tong Y, et al. Linear building pattern recognition combining Gestalt principles and convex polygon decomposition[J]. Acta Geodaetica et Cartographica Sinica, 2023, 52(1): 117-128)

[17] 郭庆胜, 魏智威, 王勇等. 特征分类与邻近图相结合的建筑物群空间分布特征提取方法[J]. 测绘学报, 2017, 46(5): 631-638. (Guo Q, Wei Z & Wang Y, et al. The method of extracting spatial distribution characteristics of buildings combined with feature classification and proximity graph[J]. Acta Geodaetica et Cartographica Sinica, 2017, 46(5): 631-638.)

[18] Wei Z, Guo Q & Wang L, et al. On the spatial distribution of buildings for map generalization[J]. Cartography and Geographic Information Science, 2018, 45(6): 539-555.

[19] He X, Zhang X & Xin Q. Recognition of building group patterns in topographic maps based on graph partitioning and random forest[J]. ISPRS Journal of Photogrammetry and Remote Sensing, 2018, 136: 26-40.

[20] Zhao R, Ai T & Yu W, et al. Recognition of building group patterns using graph convolutional network[J]. Cartography and Geographic Information Science, 2020, 47(5): 400-417.

[21] 王昊奋，漆桂林，陈华钧等. 知识图谱：方法、实践与应用[M]. 北京:电子工业出版社, 2019. (Wang H, Qi G & Chen H, et al. Knowledge Graph[M]. Beijing: Publishing House of Electronics Industry, 2019.)

[22] 蒋秉川, 万刚, 许剑等. 多源异构数据的大规模地理知识图谱构建[J]. 测绘学报, 2018, 47(8):1051-1061. (Jiang B, Wan G & Xu J, et al. Geographic knowledge graph building extracted from multi-sourced heterogeneous data[J]. Acta Geodaetica et Cartographica Sinica, 2018, 47(8): 1051-1061.)

[23] Zuheros C, Tabik S & Valdivia A, et al. Deep recurrent neural network for geographical entities disambiguation on social media data[J]. Knowledge-Based Systems, 2019, 173:117-127.

[24] 张雪英, 张春菊, 吴明光等. 顾及时空特征的地理知识图谱构建方法[J]. 中国科学: 信息科学, 2020, 50(7):1019-1032. (Zhang X, Zhang C & Wu M, et al. Spatio-temporal features based geographical knowledge graph construction. SCIENTIA SINICA Informationis, 2020, 50: 1019–1032)

[25] Neo4j Developer Guides. [OL]. https://neo4j.com/docs/cypher-manual/current/clauses/where/

[26] Yan H, Weibel R & Yang B. A multi-parameter approach to automated building grouping and generalization[J]. Geoinformatica, 2008, 12(1): 73-89.

[27] Regnauld N. Contextual building typification in automated map generalization[J]. Algorithmica, 2001, 30(2): 312-333.

第一作者简介：魏智威（1993—），男，博士，研究方向为地理信息智能化处理与可视化
First author: Wei Zhiwei(1993—)，male, Phd, majors in intelligent handling and visualization of geographical information.
E-mail: 2011301130108@whu.edu.cn

通讯作者：魏智威
Corresponding author: WEI Zhiwei
E-mail: 2011301130108@whu.edu.cn